\pdfoutput=1
\documentclass[11pt]{article}
\usepackage[margin=1in]{geometry}
\usepackage{graphicx}
\usepackage{booktabs}
\usepackage{amsmath,amssymb}
\usepackage[dvipsnames]{xcolor}
\usepackage[colorlinks=true,linkcolor=NavyBlue,citecolor=NavyBlue,urlcolor=NavyBlue]{hyperref}
\usepackage{caption}
\newcommand{\apair}{$A_{\text{pair}}$}
\newcommand{\aten}{$A_{10}$}

\title{\textbf{Can a Cacheable Decision Model Follow Rules?}}
\author{Dushyant Rajput \quad Nirdesh Chauhan \quad Siddharth Kosaraju\\[4pt]
{\normalsize AltSlate Labs LLP}\\[2pt]
{\small\texttt{dushyant@altslate.com} \quad \texttt{nirdesh@altslate.com} \quad \texttt{siddharth@altslate.com}}}
\date{September 2026}

\begin{document}
\maketitle

\begin{abstract}
A non-generative decision model scores candidate actions from their text and returns a temperature-scaled
probability, replacing a generation step with a single forward pass. The accurate way to do this reads the
state, the governing rules, and each candidate \emph{together}---a joint scorer---so every candidate is
re-encoded against every state, and cost grows with the size of the candidate menu. Independent encoding
lets each candidate be encoded once and reused across states---a measured latency reduction that widens with
the menu (about $5\times$ at $77$ candidates in our setup)---but it moves the state and the candidate apart. We ask how much \emph{rule-sensitivity}---picking the action a
stated rule requires, not the one that merely looks similar---survives that move, and whether the loss can
be trained back. On Certo, a Qwen3-4B~\cite{qwen3} decision model, we run four experiments. \textbf{(1)~At a
matched budget, the tested conversion to cacheable scoring loses rule-sensitivity: a single-vector dual encoder
falls from $1.00$ to $0.24$ recall@1 on rule-sensitive selection while the joint scorer holds $1.00$; a
shortlist-then-rerank rescue fails because the cheap encoder drops the compliant candidate before reranking.}
\textbf{(2)~Targeted supervision restores performance on held-out synthetic rule tasks}---the cacheable encoder
reads paraphrased rules ($0.99$--$1.00$), tracks counterfactual flips ($98.7\%$ of pairs correct on both sides
vs.\ $30\%$), composes operators, and partly handles rule types it never trained on, with every gain's paired
interval excluding zero and the result reproducing across seeds; we do not isolate whether these predictions
depend on the supplied rule. \textbf{(3)~On real, human-authored
rules the added benefit is not established}: after resolving a truncation confound that had masked the hard
tier, the joint scorer keeps a statistically significant edge on the short tier ($0.861$ vs.\ $0.500$; paired
CI $[+0.167,+0.556]$) and a directional, non-significant one on the hard tier ($0.655$ vs.\ $0.483$,
$n{=}29$), while the synthetic recipe gives no convincing lift over the pre-trained encoder. \textbf{(4)~A
matched cross-domain real-prose mixture did not help and reduced} unseen-source contract accuracy by $9.3$ and
$16.2$ points ($-0.093$, 95\% CI $[-0.145,-0.039]$; $-0.162$, $[-0.263,-0.062]$), though the design confounds
reduced synthetic exposure with the added prose. The picture that holds up: targeted supervision restores strong
performance across held-out synthetic rule tasks, but we do not isolate whether it makes the encoder depend on
the supplied rule, and neither it nor a cross-domain real-prose mixture demonstrably improved unseen-source
real-rule decisions; the joint scorer keeps an advantage (significant on the short tier) at the cost of caching.
Beneficial transfer to unseen-source rules is \emph{not established}; the tested privacy-data replacement
produced a measured decline whose cause and generality remain unresolved.
\end{abstract}

\section{Introduction}
Many production decisions are not open-ended generation. A support system routes a message to one of a fixed
set of intents; a workflow engine picks the one action a policy permits; a moderation layer answers a
yes/no question about a rule. For these, a large language model that writes a paragraph and then commits to
an answer is both slower than necessary and harder to calibrate than a model that simply scores the options.

Certo is a small, non-generative decision model built for this shape of problem. It encodes a
\emph{state}---the situation together with the rules that govern it---and scores each candidate action from
its text, returning a temperature-scaled probability over the candidates for three primitives: \textsc{Choice}
(pick one of several), \textsc{Score} (an ordinal level), and \textsc{Noul} (yes/no). One forward pass, no
generation, probabilities as the primary output.

The reference model uses a \emph{joint} scorer: the state, the rules, and one candidate are read together in
a single sequence, and a head emits a relevance score. Reading them together is what lets the model apply a
rule to a candidate. It is also what makes the model expensive at scale: because a candidate only ever
exists fused with a state, its representation cannot be reused, and the cost of scoring a menu grows with the
number of candidates. Serving stacks already reuse computation in a limited way: automatic prefix
caching~\cite{pagedattention} lets a causal joint scorer reuse the shared state-and-rules prefix across the
candidates of a \emph{single} state, so only each candidate's suffix is recomputed. What no joint scorer can
do is reuse a candidate's representation across \emph{different} states, because that representation is always
conditioned on the state it was scored against.

The alternative is \emph{independent encoding}, the design behind dense retrieval~\cite{dpr,sbert} and its
multi-vector variants~\cite{colbert}: encode the state once and each candidate once, then combine cheaply.
This adds the second kind of reuse---a candidate's vector is state-independent, so it is computed once and
reused across every state that shares a rubric. In our measurements this widens the latency gap with menu
size (\S\ref{sec:e4}); we report it for the implementation measured, not as an architectural constant. The
risk is equally real: the state and the candidate never meet inside the network, and rule application may be
exactly the interaction that separation destroys.

This paper is a study of that tradeoff on one model. The precise question is:

\begin{quote}
\emph{How much rule-sensitivity survives when a joint decision scorer is converted to a cacheable
independent-encoding scorer, can the loss be trained back, and does any recovered rule-sensitivity transfer
to real rules the model was not trained on?}
\end{quote}

Our contributions are four experiments and a qualified answer. Section~\ref{sec:e4} shows the conversion
loses rule-sensitivity and a shortlist rescue fails. Section~\ref{sec:ra} shows counterfactual supervision
recovers it on synthetic rules, reproducibly. Section~\ref{sec:real} tests real rules, uncovers and resolves
a truncation confound, and finds the synthetic recovery does not clearly transfer. Section~\ref{sec:pilot}
runs a controlled real-prose training pilot and finds cross-domain data does not help. We are explicit
throughout about which claims are established and which are not.

\begin{figure}[t]\centering
\includegraphics[width=\textwidth]{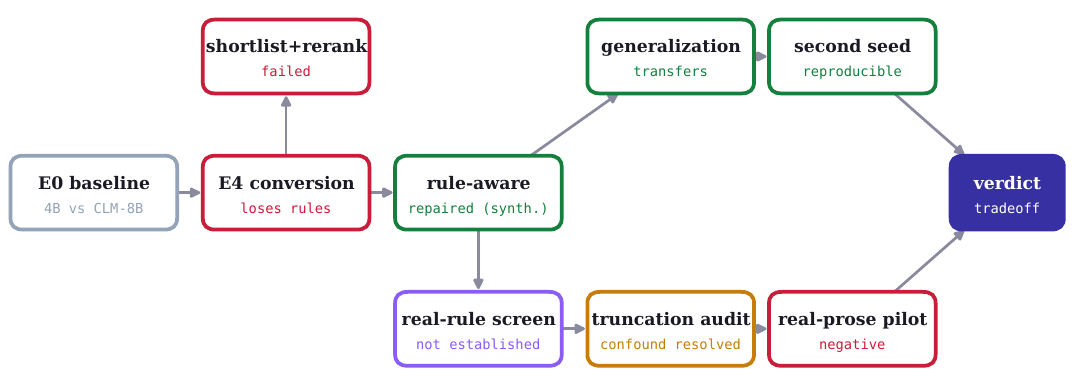}
\caption{The experimental program as a dependency graph. E0 is a prior baseline study (Certo vs.\ a contrastive
dual-encoder); E4 is the four-scorer conversion study of \S\ref{sec:e4}; later nodes are this paper's
experiments. Green nodes are positive results, red are negative, amber is a confound resolved by re-analysis.
The spine (E0\,$\to$\,E4\,$\to$\,rule-aware\,$\to$\,generalization\,$\to$\,second seed) establishes and
validates the training fix; the lower track (real-rule screen\,$\to$\,truncation audit\,$\to$\,real-prose
pilot) tests it against real rules.}
\label{fig:dag}
\end{figure}

\section{Setting and architectures}
\label{sec:arch}
\paragraph{Model.} Certo is a Qwen3-4B~\cite{qwen3} encoder. All scorers share the backbone's causal
attention and differ in how hidden states are reduced to a score (below). \textsc{Choice} and \textsc{Noul}
are trained with a multiclass Brier loss~\cite{brier}; \textsc{Score} with a ranked-probability (squared-EMD)
loss~\cite{rps}. Outputs are temperature-scaled per primitive~\cite{calibration}; we do not measure
calibration under transfer, so we describe them as temperature-scaled probabilities rather than calibrated
ones. Adapters are LoRA~\cite{lora} (rank $16$) on the attention and MLP projections.

\paragraph{Four scorers.} All read a state and a set of candidates and rank the candidates; they differ in
where the state and a candidate meet, which decides whether a candidate's representation is reusable.
\begin{itemize}\itemsep2pt
  \item \textbf{\apair{} (joint cross-encoder).} Each candidate is concatenated with the state and rules into
  one sequence; the last-token hidden state feeds a scalar-relevance head, and a softmax over candidates gives
  the distribution. Not cacheable. This is the standard reranker design~\cite{monobert}.
  \item \textbf{\aten{} (joint head).} Reads the state once; the last-token hidden state feeds a native
  $K$-way head over up to ten candidates. Accurate but capped at the head width.
  \item \textbf{B (single-vector dual).} State and each candidate are encoded separately; each one's
  last-token hidden state is linearly projected and L2-normalized to a single vector, and the score is scaled
  cosine. Candidate vectors are cacheable~\cite{dpr,sbert}.
  \item \textbf{C (multi-vector).} Each token's hidden state is linearly projected to a per-token vector (no
  pooling); the score is late-interaction MaxSim over the token vectors~\cite{colbert}. Cacheable.
\end{itemize}
Throughout, the rules live \emph{in the state}, so candidate embeddings stay reusable even for rule-sensitive
decisions: only the state carries the rubric.

\paragraph{Protocol.} ``Matched budget'' means equal training examples and optimizer steps across the arms
compared; token and compute counts vary with content and are not separately equalized. Splits are disjoint by
rule family and by document source; held-out suites use rule combinations, wordings, and types never seen in
training. Confidence intervals are bootstraps ($5000$ resamples) grouped by the unit of dependence: by
counterfactual pair for the paired block, by document elsewhere. In the real-rule evaluations each item is a
distinct document (Table~\ref{tab:real}), so document- and item-level resampling coincide there. Latencies are
p50, single query, bf16, on one GPU, with candidate vectors served from a warm cache for the cacheable arms;
they index the trend, not an optimized deployment.

\section{Experiment 1: making scoring cacheable loses rule-sensitivity}
\label{sec:e4}
\begin{figure}[t]\centering
\includegraphics[width=0.86\textwidth]{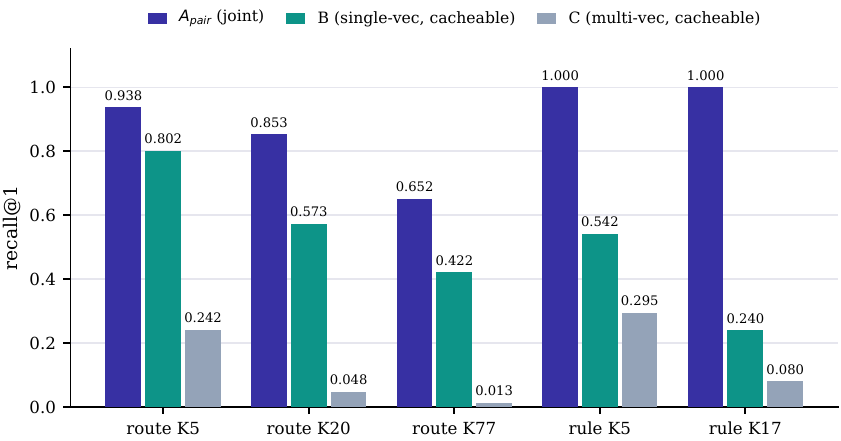}
\caption{Recall@1 by scorer. The joint scorer (\apair{}) holds rule-sensitive accuracy at $1.00$ across menu
sizes; both cacheable scorers collapse on rules (B to $0.24$, C to $0.08$ at K17) while keeping some routing
ability. The rule-sensitive slice has $17$ candidates (K17); routing K20 is a separate slice.}
\label{fig:e4}
\end{figure}

We trained all four scorers from the same reference model at a matched budget and measured recall@1 on two
kinds of decision at several menu sizes $K$: \emph{routing} (choose the intent matching a message) and
\emph{rule-sensitive} selection (choose the action a stated rule requires). Results are in
Table~\ref{tab:e4} and Figure~\ref{fig:e4}.

\begin{table}[t]\centering\small
\caption{Experiment 1: recall@1 by scorer, with p50 latency. \emph{Reusable}: candidate embeddings reusable
across states; the joint scorer can still reuse the state prefix within one state via prefix caching. The
rule-sensitive slice has $17$ candidates, so its second column is K17 (the full menu), not K20.}
\label{tab:e4}
\begin{tabular}{lccccccc}
\toprule
& \multicolumn{3}{c}{routing} & \multicolumn{2}{c}{rule-sensitive} & & cand.\ emb. \\
\cmidrule(lr){2-4}\cmidrule(lr){5-6}
scorer & K5 & K20 & K77 & K5 & K17 & latency & reusable \\
\midrule
\apair{} (joint) & 0.938 & 0.853 & 0.652 & \textbf{1.00} & \textbf{1.00} & 52--242\,ms & no \\
\aten{} (joint head) & 0.953 & --- & --- & 1.00 & --- & 47\,ms ($K{\le}10$) & no \\
B (single-vector) & 0.802 & 0.573 & 0.422 & 0.542 & 0.240 & 46\,ms & yes \\
C (multi-vector) & 0.242 & 0.048 & 0.013 & 0.295 & 0.080 & 46\,ms & yes \\
\bottomrule
\end{tabular}
\end{table}

\begin{figure}[t]\centering
\includegraphics[width=0.6\textwidth]{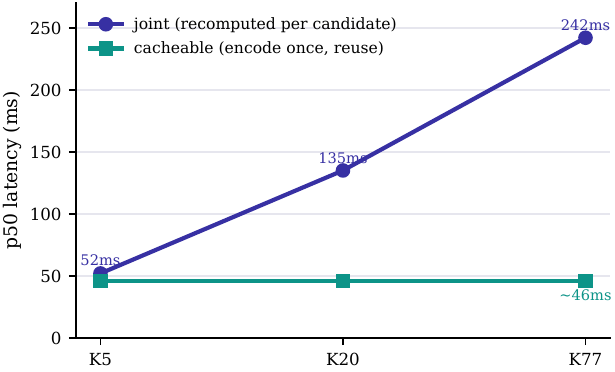}
\caption{Why caching is worth wanting. Joint scoring grows with the menu ($52\to242$\,ms, K5$\to$K77);
cacheable scoring is approximately flat ($\approx46$\,ms over the measured range) because candidate vectors are
reused. Conditions in the Protocol paragraph.}
\label{fig:lat}
\end{figure}

The joint scorer nails rule-sensitivity ($1.00$) but its latency grows with the menu (Figure~\ref{fig:lat}).
The single-vector encoder keeps partial routing but loses rule-sensitivity ($0.24$--$0.54$ vs.\ $1.00$); the
multi-vector encoder collapses to near chance, observed as a MaxSim score collapse in which no candidate
stands out. Cacheable scoring is approximately flat over the measured range (the cosine step still grows with
the candidate count, but is negligible here). The measured latency gap is about $2.9\times$ at K20 ($135$
vs.\ $46$\,ms) and $5.3\times$ at K77 ($242$ vs.\ $46$\,ms), the largest menu we measure, and widens with $K$;
these are execution latencies, not scoring-pass counts (Figure~\ref{fig:lat}).

\paragraph{A shortlist rescue also fails.} We tried a hybrid: retrieve a cheap top-$k$ with the cacheable
encoder, then rerank with the joint scorer. The target is candidate-inclusion recall@$k$---the rule-compliant
answer appearing in the top-$k$---at $\ge0.98$, so a compute-saving $k$ (smaller than the menu) still leaves
the reranker a correct option to pick. The rule-sensitive slice has $17$ candidates (the K17
column of Table~\ref{tab:e4}). At the tested shortlist sizes, inclusion recall was $0.72$/$0.80$ at
$k{=}5$/$10$ and reached $1.00$ only at $k{=}17$, i.e.\ the full set (no shortlist); $k{=}11$--$16$ were not
measured. On routing, inclusion recall was $0.70$/$0.82$/$0.93$ at $k{=}5$/$10$/$20$ and hybrid top-1 accuracy
($0.57$--$0.64$) did not exceed the joint reranker's $0.65$. So for the tested reduced shortlists, no size both
saved compute and met the target. The omitted candidates were the rule-compliant ones, which is consistent with
surface-similarity ranking but does not by itself establish that as the cause. \textbf{Among the conversions
tried at this budget, only joint interaction preserved rule-sensitive accuracy.}

\section{Experiment 2: counterfactual supervision recovers it on synthetic rules}
\label{sec:ra}
Rather than change the architecture, we changed the data. The single-vector encoder (B) was continued on
\emph{counterfactual} rule pairs: two situations that differ by exactly one condition---an exception firing,
a threshold crossed---so the correct action flips. The rule is written into the state, keeping candidates
cacheable, and the semantically tempting but prohibited action is always present as a hard
negative~\cite{cad}. Positives are labelled by an executable rule.

\paragraph{Example.} A counterfactual pair, differing in one clause (the candidate menu is identical):
\emph{``Policy: to release the payment, do \texttt{release\_payment}. Exception: if the request is past the
deadline, do \texttt{deny\_out\_of\_window}. Situation: the request is \underline{within} the deadline.''}
$\to$ gold \texttt{release\_payment}; flipping the last clause to \underline{past} the deadline moves the gold
to \texttt{deny\_out\_of\_window}, with \texttt{release\_payment} now the hard negative.

\begin{figure}[t]\centering
\includegraphics[width=0.9\textwidth]{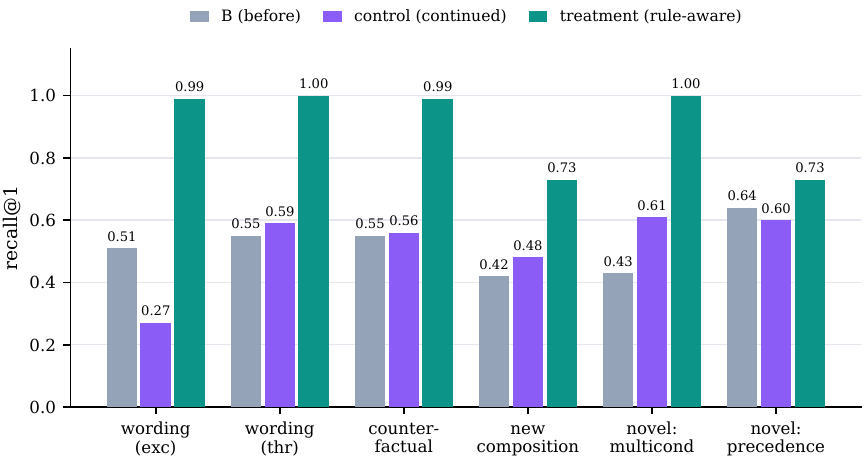}
\caption{Held-out generalization suite. The treatment reads paraphrased rules ($0.99$--$1.00$), tracks
counterfactuals, composes an exception with a threshold ($0.73$), and partly handles untrained rule types
(multi-condition $1.00$, precedence $0.73$), with no statistically detectable routing change (the interval
still admits an $\approx8$-point drop, so this is not an equivalence claim).}
\label{fig:gen}
\end{figure}

Evaluation used a frozen suite of held-out rules across four blocks (paraphrased wordings, counterfactual
pairs, a composition of two operators, and rule types never trained). Table~\ref{tab:gen} and
Figure~\ref{fig:gen} report recall@1 with paired bootstrap intervals against two baselines: the encoder
before this training (B) and a control continued on unrelated data.

\begin{table}[t]\centering\small
\caption{Experiment 2: held-out generalization suite, recall@1. Slashes give the two families within a block;
the CI is for the pooled block. Bootstrap grouping: by pair for the counterfactual block ($300$ items, $150$
pairs), by item otherwise ($n{=}300$). Every block gain over both baselines excludes zero.}
\label{tab:gen}
\begin{tabular}{lcccc}
\toprule
block & B (before) & control & treatment & treat.\ $-$ control (95\% CI) \\
\midrule
unseen wording (exc.\,/\,thr.), $n{=}300$ & 0.51\,/\,0.55 & 0.27\,/\,0.59 & \textbf{0.99\,/\,1.00} & $+0.563\ [0.507,0.620]$ \\
counterfactual, $150$ pairs & 0.55 & 0.56 & \textbf{0.99} & $+0.430\ [0.377,0.483]$ \\
new composition, $n{=}300$ & 0.42 & 0.48 & \textbf{0.73} & $+0.250\ [0.187,0.313]$ \\
novel type (multicond.\,/\,prec.), $n{=}300$ & 0.43\,/\,0.64 & 0.61\,/\,0.60 & \textbf{1.00\,/\,0.73} & $+0.263\ [0.217,0.313]$ \\
\midrule
counterfactual both-correct & 0.30 & 0.29 & \textbf{0.987} & --- \\
routing (retention) & 0.430 & 0.453 & 0.420 & $-0.033\ [-0.077,+0.013]$ \\
\bottomrule
\end{tabular}
\end{table}

The treatment reads paraphrased rules it never saw, answers both sides of a counterfactual pair correctly
$98.7\%$ of the time against $30\%$ for the baseline, composes operators, and partly handles rule types
outside its training. We detect no statistically significant routing change ($-0.033$, 95\% CI
$[-0.077,+0.013]$); the interval still admits up to a roughly $8$-point drop, so this is not an equivalence
claim. The result reproduced under a second seed (seed 2 recall@1: wording $1.00$/$1.00$, counterfactual
$1.00$, composition $0.82$, multi-condition $1.00$, precedence $0.73$; both-correct $1.00$), with the
precedence ceiling identical across seeds. The cacheable encoder therefore \emph{generalizes across the
held-out synthetic rule tasks} while keeping the large-menu serving advantage. We stop short of calling it
rule-following: whether its predictions depend on the supplied rule, rather than on a learned decision pattern,
is what the rule-only control (below) would settle, and we do not claim it here.

\paragraph{What the synthetic suite does not yet isolate.} These pairs change a condition in the state, which
shows sensitivity to the relevant facts but does not by itself rule out a learned, template-specific
procedure. The decisive control---holding the facts and candidates fixed and changing only the stated rule, so
the correct answer moves, with deleted-rule and shuffled-rule baselines---is not yet run (\S\ref{sec:future}).
The other open question is external validity: the suite is synthetic and templated.

\section{Experiment 3: real rules, and a truncation confound}
\label{sec:real}
We scored the trained encoders on real, human-authored decisions drawn from JevBench, an internal benchmark
of insurance, policy, and contract questions with the rules stated in the text. A first pass read as a flat
null on the hard tier. That reading was an artifact of truncation.

\begin{figure}[t]\centering
\includegraphics[width=0.7\textwidth]{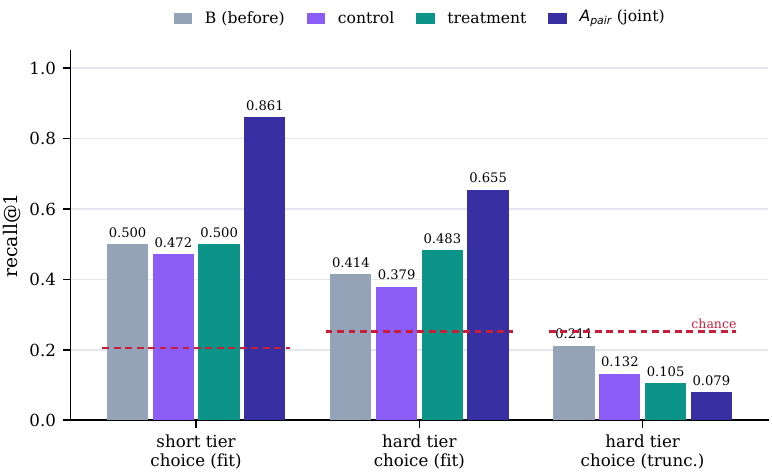}
\caption{Real rule-sensitive choice within the trained window (dashed red: chance). The joint scorer keeps an
advantage on inputs that fit (significant on the short tier); on truncated inputs every scorer falls to near or
below chance, the joint one furthest, because right-truncation drops its appended candidate.}
\label{fig:real}
\end{figure}

\paragraph{The confound.} The joint scorer appends the candidate after the state in one sequence. On a long
document, right-truncation at the trained window cuts the candidate off entirely. An input audit found that
only $29$ of $67$ hard-tier choice items fit both scorers' trained windows (state up to $3{,}649$ tokens
against a $640$/$704$ budget); the short tier fit in full. Rescoring only within the trained window, with
the models frozen, separates skill from truncation (Table~\ref{tab:real}, Figure~\ref{fig:real}).

\begin{table}[t]\centering\small
\caption{Experiment 3: real rule-sensitive choice, recall@1, rescored within the trained window. On truncated
items the joint scorer falls furthest, exactly as the candidate-dropping mechanism predicts.}
\label{tab:real}
\begin{tabular}{lccccc}
\toprule
slice & $n$ & B & control & treatment & \apair{} (joint) \\
\midrule
short tier, choice (fits) & 36 & 0.500 & 0.472 & 0.500 & \textbf{0.861} \\
hard tier, choice (fits) & 29 & 0.414 & 0.379 & 0.483 & \textbf{0.655} \\
hard tier, choice (truncated) & 38 & 0.211 & 0.132 & 0.105 & 0.079 \\
short tier, yes/no (fits) & 24 & 0.958 & 0.958 & 0.917 & 0.958 \\
\bottomrule
\end{tabular}
\end{table}

Two conclusions, held apart, with paired bootstraps over the fitting items. On the short tier the joint scorer
has a significant edge over the cacheable treatment ($0.861$ vs.\ $0.500$, i.e.\ $31/36$ vs.\ $18/36$; paired
$d=+0.361$, 95\% CI $[+0.167,+0.556]$). On the hard tier the edge is directional but \emph{not} significant
($0.655$ vs.\ $0.483$, i.e.\ $19/29$ vs.\ $14/29$; $d=+0.172$, CI $[-0.069,+0.414]$)---a five-item difference
on a small slice, which we report descriptively and treat as weaker than the short-tier result. The $38$
truncated hard items are set aside, not repaired, so the shift from a raw $0.328$ to $0.655$ is a subset
comparison, not a measured gain on all $67$. Separately, the synthetic training gave no convincing added
benefit over the pre-trained encoder on these real rules (treatment vs.\ B: short $d=0.000$, CI
$[-0.167,+0.167]$; hard $d=+0.069$, CI $[0.000,+0.172]$). Transfer of the synthetic recovery to real rules is
\emph{not established}, which is weaker than saying it was disproved.

\section{Experiment 4: a controlled real-prose pilot}
\label{sec:pilot}
\begin{figure}[t]\centering
\includegraphics[width=0.82\textwidth]{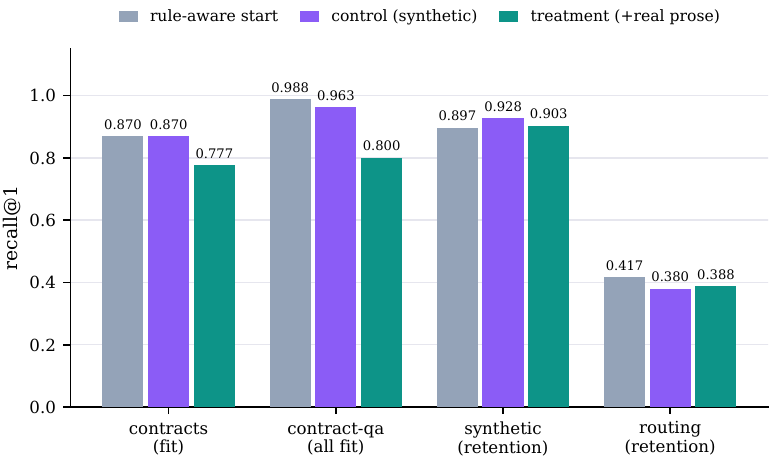}\\[6pt]
\includegraphics[width=0.78\textwidth]{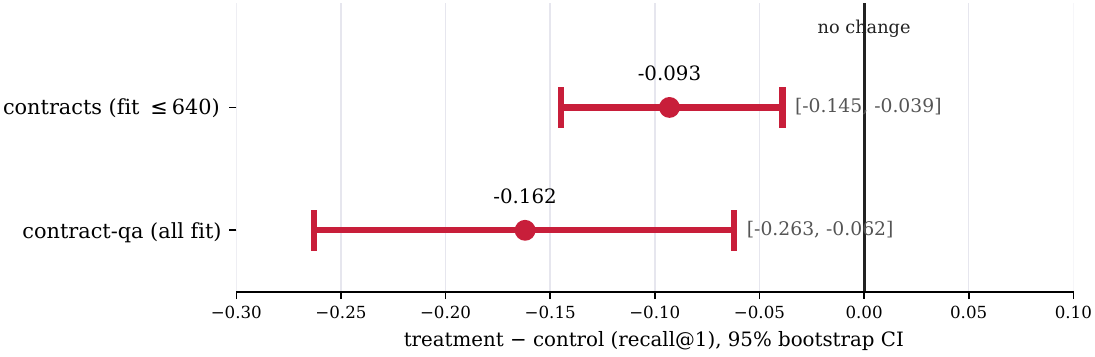}
\caption{Top: pilot recall@1. On both contract tests the treatment falls below the control and the untrained
start; retention on synthetic rules and routing is approximately unchanged (no equivalence test). Bottom:
treatment $-$ control with 95\% bootstrap intervals, both fully left of zero.}
\label{fig:pilot}
\end{figure}

The natural next intervention is to train on real rule \emph{prose} rather than synthetic templates. We ran
it as a controlled pilot with a matched continuation. Both arms started from the rule-aware encoder with the
same allowance ($6{,}000$ examples, $500$ steps, one seed, identical replay). The control continued the
synthetic mixture; the treatment replaced $40\%$ of the synthetic slot with real privacy-policy decisions
(LegalBench~\cite{legalbench} \texttt{privacy\_policy\_entailment}, rules in the state, yes/no). The locked
primary test was a source-disjoint corpus---contract questions (\texttt{consumer\_contracts\_qa} and
\texttt{contract\_qa})---never seen in training. We report inputs that fit the trained window separately from
those that do not, per an input-validity check. Results are in Table~\ref{tab:pilot} and
Figure~\ref{fig:pilot}.

\begin{table}[t]\centering\small
\caption{Experiment 4: real-prose pilot, recall@1. Replacing part of the synthetic mixture with privacy-policy
prose reduced unseen-source contract accuracy ($-9.3$ and $-16.2$ points); retention on synthetic rules and
routing is approximately unchanged (we do not run an equivalence test).}
\label{tab:pilot}
\begin{tabular}{lcccc}
\toprule
slice & $n$ & rule-aware start & control (synthetic) & treatment (+real prose) \\
\midrule
contracts (fit $\leq$640) & 332 & 0.870 & 0.870 & \textbf{0.777} \\
contract-qa (all fit) & 80 & 0.988 & 0.963 & \textbf{0.800} \\
contracts (truncated) & 144 & 0.569 & 0.562 & 0.556 \\
synthetic rules (retention) & 1200 & 0.897 & 0.928 & 0.903 \\
routing (retention) & 400 & 0.417 & 0.380 & 0.388 \\
\bottomrule
\end{tabular}
\end{table}

Both contract differences sit fully to the left of zero: $-0.093$ (CI $[-0.145,-0.039]$, a $9.3$-point drop)
on the fitting slice and $-0.162$ (CI $[-0.263,-0.062]$) on \texttt{contract-qa}. The supported conclusion is
narrow: \emph{replacing part of the synthetic mixture with privacy-policy examples reduced contract accuracy
under this recipe}. We cannot separate the causes, because the treatment changes two things at once---it
removes synthetic examples \emph{and} adds privacy ones---so the harm could come from reduced synthetic
exposure, the added prose, or their interaction. The control matching the untrained start on contracts
($0.870=0.870$) rules out neither. A third arm with the same reduced synthetic exposure but no privacy
supervision would isolate this (\S\ref{sec:future}). Under the pre-registered rule---advance only on a
meaningful gain over continuation---the pilot stops here, with no second seed. Its scope is one source pairing
(privacy $\to$ contracts) in a binary format that does not exercise the joint-vs-cacheable choice gap; it
answers the pilot's exact question for this pairing: no.

\section{Discussion}
The four experiments compose into a single, qualified reading. A cacheable independent-encoding decision model
\emph{generalizes across held-out variants of its training distribution} (paraphrase, counterfactual,
composition, a second seed); whether it does so by interpreting the supplied rule, rather than a learned
decision pattern, is what the rule-only control would settle, and we do not claim it here. What did not hold is
transfer: neither the synthetic recipe (Experiment~3) nor a cross-domain real-prose mixture (Experiment~4)
demonstrably improved decisions on unseen-source real rules. Meanwhile the joint scorer wins significantly on
the short real-rule slice (the hard-tier edge is uncertain), and pays for it with candidate recomputation.

\paragraph{Practical implication.} The result is an application-dependent tradeoff, not a general
recommendation. Where menus are small and rule-sensitivity matters, the joint scorer is the safer choice; we
report its latency (Table~\ref{tab:e4}) and leave the affordability judgment to the application. Where menus are
large, the cacheable encoder's serving advantage is real, but so is its accuracy cost even on
routing---$0.422$ vs.\ $0.652$ for the joint scorer at K77---so the choice depends on how much accuracy the
application can trade for latency. A promising but untested hypothesis is that training on the customer's own
decision types is what makes Certo strong; we did not demonstrate this on a customer dataset or a source-matched
real-rule experiment, and we did not establish that either scorer generalizes to unseen-source rules for free.

\paragraph{A methodological note.} Experiment~3 is a reminder that an evaluation window is part of the model.
The hard-tier ``null'' was an artifact of appending the candidate after a long document and truncating from
the right; the fix was an input audit and a within-window rescore, not more training. Reporting fitting and
truncated inputs separately is cheap insurance against this class of confound.

\section{Future work}
\label{sec:future}
Several experiments would sharpen or overturn the negative-transfer reading, roughly in order of expected
value.

\begin{enumerate}\itemsep3pt
  \item \textbf{Rule-only counterfactual test.} The decisive control for rule use: hold the facts and
  candidates fixed and change only the stated rule so the correct answer moves, with deleted-rule and
  shuffled-rule baselines. This would separate genuine rule application from a learned, template-specific
  procedure, which the current condition-flipping pairs do not fully isolate.
  \item \textbf{A third pilot arm.} To separate reduced synthetic exposure from added privacy prose in
  Experiment~4, add an arm with the same reduced synthetic count and no privacy supervision.
  \item \textbf{Choice-format real rules.} Experiments~3--4 lean on binary and short-clause real tasks; the
  joint-vs-cacheable gap is clearest on multi-candidate \textsc{Choice}. A real, multi-candidate rule corpus
  (for example contract-clause selection or statutory reasoning framed as choice) would test the gap where
  it matters most.
  \item \textbf{A prefix-reusing joint latency baseline.} Our latency comparison is against a joint scorer
  without prefix caching. A joint scorer that reuses the shared state prefix across candidates would narrow the
  gap; measuring it would scope the caching advantage precisely.
  \item \textbf{Source-matched training.} The pilot trained on privacy policies and tested on contracts. A
  matched pilot---train and test on the same document family, split by document---would separate ``real prose
  does not help'' from ``cross-domain real prose does not help.''
  \item \textbf{A joint model on the same new data.} To attribute the remaining real-rule gap specifically to
  architecture, the joint scorer should be trained on the same real corpus as the cacheable one; only then is
  the comparison an architecture comparison rather than a data comparison.
  \item \textbf{Longer windows and centre-preserving truncation.} The joint scorer's real-rule ceiling is
  entangled with context length. Training and serving at longer windows, and truncating the state rather than
  the appended candidate, would remove the confound of Experiment~3 by construction.
  \item \textbf{The multi-vector collapse.} C's failure was observed as a MaxSim score collapse but its
  mechanism (long causal state swamping late interaction) is proposed, not isolated; a length/mask/aggregation
  ablation would settle it, and a rule-aware multi-vector encoder may recover more than the single-vector one.
  \item \textbf{The precedence ceiling.} Recall@1 on the untrained precedence rule type tops out at $0.73$ on
  both seeds; targeted supervision for ordered-rule application is the obvious next lever.
  \item \textbf{Calibration under transfer.} We report accuracy under transfer but not calibration; whether a
  rule-aware cacheable encoder stays calibrated on unseen-source rules is a separate, deployment-relevant
  question.
\end{enumerate}

\section{Limitations}
The real-rule evaluations are small (tens of items per slice) with wide intervals, and most are binary or
short-clause; the joint-vs-cacheable choice gap rests on the JevBench choice slices. The rule-aware result
establishes generalization across held-out synthetic rule tasks, not that predictions depend on the supplied
rule; the rule-only control that would isolate this (\S\ref{sec:future}) is not yet run. The real-prose pilot
is a single source pairing and a single seed;
a different pairing or a choice-format corpus could differ. C's collapse mechanism is proposed, not isolated.
Attributing the real-rule gap specifically to architecture, rather than to data, would require the joint
model trained on the same new corpus, which we did not run. All results are on one backbone (Qwen3-4B) and
one adapter recipe.

\section{Reproducibility}
All experiments run on NVIDIA RTX PRO 4500 Blackwell Server Edition GPUs (four for training, one for the
latency measurements). Adapters are LoRA (rank $16$, $\alpha{=}32$) on the \texttt{q,k,v,o,gate,up,down}
projections, AdamW at learning rate $1\times10^{-4}$ (cosine, $100$ warmup steps), bf16, gradient
checkpointing. \textbf{Experiment~1} trains all four arms for $1000$ steps from the same reference model with
effective batch $64$ (matched budget): the dual arm at state length $640$, candidate length $64$, projection
$512$ (per-device batch $8$, accumulation $2$); the joint cross-encoder \apair{} at sequence length $704$
(per-device batch $4$, accumulation $4$); the multi-vector arm at state length $512$, candidate length $48$,
projection $128$; \aten{} is the native joint head of the reference model. \textbf{Experiment~2} continues the
dual arm for $600$ steps (control on a neutral mixture, treatment on the counterfactual mixture; both from the
Experiment~1 dual checkpoint), and \textbf{Experiment~4} for $500$ steps, each with the same optimizer settings.
Checkpoints for the cacheable arms are on Hugging Face (\texttt{rajpdus/certo-rule-aware-arms}). The scripts for
all four experiments, the frozen generalization suite, the locked real-prose test set, and per-item prediction
files---from which every reported interval can be recomputed without retraining---are released at
\url{https://github.com/AltSlate-Labs/certo-rules-or-reuse} (tag \texttt{v1.0}, commit \texttt{c0c0472}). Every
real-rule and pilot evaluation reports inputs that fit the trained window separately from those that do not;
treatment-vs-control differences are bootstraps grouped by the unit of dependence (\S\ref{sec:arch}).

\bibliographystyle{unsrt}
\bibliography{refs}

\end{document}